\pdfoutput=1

\documentclass[10pt, a4paper]{article}

\usepackage[final]{lrec2026}

\usepackage{adjustbox}
\usepackage{amsmath,amssymb}
\usepackage[edges]{forest}
\usepackage{booktabs,tabularx}
\usepackage{placeins}
\usepackage{afterpage}
\usepackage{dblfloatfix}%
\usepackage[table]{xcolor}%

\usepackage{multirow}
\usepackage{hyperref}%
\usepackage{array}%
\usepackage{float}%
\usepackage{subcaption}%
\usepackage{xspace}
\usepackage{url}
\usepackage{ragged2e}%
\usepackage{comment}
\usepackage{threeparttable}

\usepackage{booktabs,siunitx,xcolor}
\newcommand{\Ta}{\ensuremath{T_{\!A}}}
\newcommand{\Tt}{\ensuremath{T_{\!T}}}
\newcommand{\Ti}{\ensuremath{T_{\!I}}}
\newcommand{\Tf}{\ensuremath{T_{\!\text{ final}}}}

\newcommand{\ta}{\ensuremath{\theta_{\!\text{ align}}\xspace}} 

\newcommand{\Ocat}{\ensuremath{\mathcal{O}}\xspace}
\newcommand{\Tcat}{\ensuremath{\mathcal{T}}\xspace}
\newcommand{\TNOcat}{\ensuremath{\mathcal{TNO}}\xspace}
\newcommand{\TOcat}{\ensuremath{\mathcal{TO}}\xspace}
\newcommand{\TOAcat}{\ensuremath{\mathcal{TOA}}\xspace}

\newcommand{\Oold}{\ensuremath{\mathcal{O}_{\!\text{ old}}\xspace}}
\newcommand{\Orec}{\ensuremath{\mathcal{O}_{\!\text{ recent}}\xspace}}

\newcommand{\Tfirst}{\ensuremath{\mathcal{T}_{\!\text{ first}}\xspace}}
\newcommand{\Tlate}{\ensuremath{\mathcal{T}_{\!\text{ late}}\xspace}}

\newcommand{\TOAfirst}{\ensuremath{\mathcal{TOA}_{\!\text{ first}}}\xspace}
\newcommand{\TOAlate}{\ensuremath{\mathcal{TOA}_{\!\text{ late}}}\xspace}

\newcommand{\TODlate}{\ensuremath{\mathcal{TOD}_{\!\text{ late}}}\xspace}

\newcommand{\HN}{\textsc{HydroNewsFr}\xspace}

\newcommand{\td}{\ensuremath{\theta_{\!\text{ delay}}\xspace}}%
\newcommand{\Delay}{\ensuremath{\Delta T}}%

\definecolor{ltgray}{gray}{0.92}

\newcommand{\Lone}{\hspace*{1.0em}}
\newcommand{\Ltwo}{\hspace*{2.0em}}
\newcommand{\Lthr}{\hspace*{3.0em}}

\usepackage{pifont}
\newcommand{\cmark}{\ding{51}}%
\newcommand{\xmark}{\ding{55}}%

\usetikzlibrary{patterns.meta,arrows.meta,positioning}

\tikzset{
  topicbar/.style={
    fill=gray!105,
    draw=gray!50
  },
  topicpattern/.style={
    pattern = {Lines[angle=45, distance=5pt, line width=0.25pt]},
    pattern color = gray!70
  }
}

\tikzset{
  outlierbar/.style={
    fill=gray!15,
    draw=gray!15
  },
  outlierpattern/.style={
    pattern = {Lines[angle=45, distance=6pt, line width=0.25pt]},
    pattern color = gray!50
  }
}

\tikzset{
  tdot/.style={
    circle,
    inner sep=0pt,
    minimum size=2.2pt,
    fill=black
  },
  idot/.style={
    circle,
    inner sep=0pt,
    minimum size=2.6pt,
    fill=blue
  }
}

\newcommand{\TickUp}{0.32}
\newcommand{\timelineaxis}{\draw[->] (0,0) -- (5.5,0) node[pos=0,left]{\(t\)};}

\tikzset{labelbg/.style={fill=white, inner sep=1pt, rounded corners=0.8pt, text=black}}

\tikzset{
  tdot/.style={circle,inner sep=0pt,minimum size=2.2pt,fill=black},
  idot/.style={circle,inner sep=0pt,minimum size=2.6pt,fill=black},%
}

\newcommand{\topicbelt}[2]{%
  \fill[topicbar]     (#1,0.38)  rectangle (#2,0.68);%
  \fill[topicpattern] (#1,0.38)  rectangle (#2,0.68);%
}
\newcommand{\outlierbelt}[2]{%
  \fill[outlierbar]     (#1,-0.68) rectangle (#2,-0.38);
  \fill[outlierpattern] (#1,-0.68) rectangle (#2,-0.38);
}

\tikzset{
  seededge/.style={ultra thick},%
}
\forestset{
  seeder/.style={fill=gray!18},%
}

\title{From Noise to Signal: When Outliers Seed New Topics}

\name{
\begin{tabular}{c}
\bfseries Evangelia Zve$^{1,2}$, Gauvain Bourgne$^{1}$, Benjamin Icard$^{1}$,\\
and Jean-Gabriel Ganascia$^{1}$\\[0.03in]
\end{tabular}
}

\address{
$^{1}$ Sorbonne Université, CNRS, LIP6, France\\
$^{2}$ Infopro Digital, France\\[0.03in]
\{evangelia.zve, gauvain.bourgne, benjamin.icard, jean-gabriel.ganascia\}@lip6.fr\\
evangelia.zve@infopro-digital.com
}

\abstract{
Outliers in dynamic topic modeling are typically treated as noise, yet we show that some can serve as early signals of emerging topics. We introduce a \emph{temporal taxonomy} of news-document trajectories that defines how documents relate to topic formation over time. It distinguishes \emph{anticipatory outliers}, which precede the topics they later join, from documents that either reinforce existing topics or remain isolated. By capturing these trajectories, the taxonomy links weak-signal detection with temporal topic modeling and clarifies how individual articles anticipate, initiate, or drift within evolving clusters. We implement it in a cumulative clustering setting using document embeddings from eleven state-of-the-art language models and evaluate it retrospectively on \HN{}, a French news corpus on the hydrogen economy. Inter-model agreement reveals a small, high-consensus subset of anticipatory outliers, increasing confidence in these labels. Qualitative case studies further illustrate these trajectories through concrete topic developments.
\\ \newline
}

\begin{document}
\pagestyle{empty}

\maketitleabstract

\section{Introduction}

Anticipating emerging topics in fast-evolving news streams is essential for tracking public debate, identifying opportunities, and monitoring risks~\cite{ansoff1980weak}. However, most topic-modeling methods identify topics only after coherent clusters have formed, which limits their ability to capture early signals~\cite{churchill2022evolution}.

This limitation stems in part from how these methods handle atypical documents~\cite{zve-etal-2025-outliers}. Such documents often appear before a topic stabilizes, yet mainstream approaches typically assign them to diffuse clusters or discard them as noise~\cite{hiltunen2008future}. We argue that topic modeling should account for the temporal relationship between documents and evolving clusters, while also distinguishing \emph{outliers} that may signal emerging topics. In practice, this requires clustering-based topic-modeling methods~\cite{grootendorst2022bertopic} that can infer the number of topics, align topic clusters over time, and explicitly account for anomalies.

In this paper, we propose a document-level perspective on topic emergence. We introduce a \emph{temporal taxonomy} of news-document trajectories that characterizes how documents relate to topic formation over time through three key events: a document's first appearance, the emergence of a topic cluster, and, when relevant, the document's first integration into that cluster. Within this framework, we define \emph{anticipatory outliers} as documents that precede the topic clusters they later join.

To validate the taxonomy, we apply it retrospectively to \HN{}, a curated French news corpus on the hydrogen economy. Articles are embedded with eleven text-embedding models and clustered cumulatively in daily windows, after which the resulting topic clusters are aligned over time. Documents are then labeled according to the taxonomy based on their observed trajectories. We evaluate the robustness of these labels through inter-model agreement and complement this analysis with qualitative case studies.

Section~\ref{sec:related_work} reviews prior work on topic modeling and weak-signal detection.
Section~\ref{sec:taxonomy} introduces the taxonomy.
Section~\ref{sec:dataset} describes the corpus, and Section~\ref{sec:framework} details the modeling framework.
Sections~\ref{sec:results} and~\ref{sec:case-studies} present the quantitative results and representative case studies.
Section~\ref{sec:conclusion} concludes and outlines directions for future work.

\section{Related Work}
\label{sec:related_work}

Dynamic topic modeling has been applied in diverse domains, from policy discourse~\cite{wang2006topics} and corporate strategy~\cite{kaplan2015double} to scientific research~\cite{hall2008history}. It builds on probabilistic methods such as Latent Dirichlet Allocation (LDA)~\cite{blei2003latent}, which infers topics from word co-occurrence patterns. Temporal extensions, notably Dynamic Topic Models (DTM)~\cite{blei2006dynamic} and the Dynamic Embedded Topic Model (DETM)~\cite{dieng2019dynamic}, a neural variant, incorporate time-awareness into the inference process. Unlike static topic models, these methods capture how topics evolve over time. A key limitation is that they require the number of topics to be specified in advance~\cite{chuang2013topic}. They also assign every document to a topic, which obscures outliers.

More recent embedding-based methods such as \texttt{BERTopic}~\cite{grootendorst2022bertopic} use document representations from pretrained language models and identify topics by clustering in the resulting semantic space. Their temporal extensions~\cite{boutaleb-etal-2024-bertrend} approximate topic evolution through post-hoc alignment of static clusters across time windows, rather than by modeling temporal dependencies directly. Our framework follows this strategy by aligning cumulative windows (Section~\ref{sec:topic_alignment}). The results of these methods are sensitive to the choice of clustering algorithm. Partition-based algorithms such as \texttt{KMeans}~\cite{hartigan1979kmeans} require a preset number of clusters and force all documents into topics, whereas density-based methods such as \texttt{HDBSCAN}~\cite{mcinnes2017hdbscan} and \texttt{OPTICS}~\cite{ankerst1999optics} can infer the number of clusters and label low-density documents as outliers. Typically discarded as noise, these outliers may instead indicate emerging topics, an aspect that remains largely overlooked in dynamic topic modeling~\cite{zve-etal-2025-outliers}.

Research on computational early detection in text streams spans multiple traditions. Event and first-story detection~\cite{allan2002introduction} focus on short-term novelty by identifying events as they appear. While our approach also targets early signals at the document level, it is informed by weak-signal analysis~\cite{hiltunen2008future}, which seeks subtle, fragmented cues that may foreshadow broader thematic shifts. Subsequent work~\cite{christophe2021monitoring} extends this approach to large corpora, but does not connect these signals to topic formation. 

To address this gap, we introduce a taxonomy of document trajectories that characterizes anticipatory outliers as weak signals of emerging topics.

\section{Taxonomy of News Trajectories}
\label{sec:taxonomy}

We introduce a \emph{temporal taxonomy} to characterize how individual documents evolve with respect to topic formation. 
Inspired by Allen's interval algebra for temporal reasoning~\cite{allen1983maintaining}, we represent each document through three key events: its first appearance (\Ta), the creation of its eventual topic (\Tt), and its first integration into that topic (\Ti), summarized in Table~\ref{tab:notation}.
Comparing their relative order yields distinct temporal behaviors, including \emph{anticipatory outliers} (\(\Ta < \Tt \le \Ti\)), in which a document appears before the topic it later joins.

\begin{table}[H]
\centering
\scriptsize
\begin{tabularx}{0.95\columnwidth}{lX}
\toprule
Symbol & Definition \\
\midrule
\(\Ta\) & Document appearance time in the stream \\
\(\Tt\) & Topic creation time \\
\(\Ti\) & Document integration time \\
\(\Tf\) & Final time window in the corpus \\
\(\td\) & Delay cutoff separating persistent and recent outliers \\
\bottomrule
\end{tabularx}
\caption{Temporal notation used in the taxonomy}
\label{tab:notation}
\end{table}

The taxonomy summarized in Table~\ref{tab:cases-symbolic} partitions documents into \emph{seven} mutually exclusive cases determined by the order and presence of these temporal events, as shown in the decision tree in Figure~\ref{fig:tree-cases-equations}. 
The symbolic notation (\ensuremath{\mathcal{T}}, \ensuremath{\mathcal{O}}, \ensuremath{\mathcal{A}}, \ensuremath{\mathcal{D}}, \ensuremath{\mathcal{NO}}), derived from the initials of \textit{Topic}, \textit{Outlier}, \textit{Anticipatory}, \textit{Drift}, and \textit{Non-Outlier}, provides an intuitive shorthand for each document's trajectory type.

\begin{table}[H]
\centering
\scriptsize
\setlength{\tabcolsep}{4pt}
\renewcommand{\arraystretch}{1.08}

\newcolumntype{L}{>{\RaggedRight\arraybackslash}X}
\newcolumntype{C}{>{\centering\arraybackslash}c}
\newcommand{\grouprowIII}[2]{%
  \addlinespace[0.6ex]%
  \multicolumn{3}{@{}l}{#1%
    \if\relax\detokenize{#2}\relax\else\ \textit{\scriptsize(#2)}\fi}%
  \\[-0.2ex]\addlinespace[0.2ex]}
\newcommand{\caserowIII}[3]{#1 & #2 & #3 \\}
\newcommand{\groupwithcondIII}[3]{%
  \addlinespace[0.6ex]%
  \multicolumn{2}{@{}l}{#1\ #2} & #3\\[-0.2ex]\addlinespace[0.2ex]}

\begin{tabularx}{\columnwidth}{@{}l L C@{}}
\toprule
Symbol & Description & Condition \\
\midrule

\grouprowIII{\(\Tcat\)}{integrated into \ensuremath{\mathcal{T}}opic}

\grouprowIII{\Lone \(\TOcat\)}{\ensuremath{\mathcal{O}}utlier integrated into \ensuremath{\mathcal{T}}opic}
\grouprowIII{\Ltwo \(\TOAcat\)}{\ensuremath{\mathcal{A}}nticipatory \ensuremath{\mathcal{O}}utlier integrated into \ensuremath{\mathcal{T}}opic}

\rowcolor{gray!10}%
\caserowIII{\Lthr \TOAfirst}{at topic creation}{\(\Ta<\Ti=\Tt\)}

\rowcolor{gray!10}%
\caserowIII{\Lthr \TOAlate}{after topic creation}{\(\Ta<\Tt<\Ti\)}

\groupwithcondIII{\Ltwo \TODlate}{(\textit{\ensuremath{\mathcal{D}}rift \ensuremath{\mathcal{O}}utlier integrated into \ensuremath{\mathcal{T}}opic})}{\(\Tt<\Ta<\Ti\)}

\grouprowIII{\Lone \(\TNOcat\)}{\ensuremath{\mathcal{N}}on-\ensuremath{\mathcal{O}}utlier integrated into \ensuremath{\mathcal{T}}opic}
\caserowIII{\Ltwo \Tfirst}{at topic creation}{\(\Ta=\Ti=\Tt\)}
\caserowIII{\Ltwo \Tlate}{after topic creation}{\(\Tt<\Ta=\Ti\)}

\addlinespace[0.6ex]
\grouprowIII{\(\Ocat\)}{\ensuremath{\mathcal{O}}utlier, not integrated into a topic}
\caserowIII{\Lone \Orec}{appeared recently}{\(\Tf-\Ta<\td\)}
\caserowIII{\Lone \Oold}{long-standing outlier}{\(\Tf-\Ta\ge\td\)}

\bottomrule
\end{tabularx}
\caption{Formal conditions for the taxonomy cases}
\label{tab:cases-symbolic}
\end{table}

Two cases correspond to documents that never integrate into a topic during the study period.  
They are distinguished by their recency with respect to the corpus end time (\Tf): recently isolated documents are labeled \Orec\ and their eventual trajectory may extend beyond \Tf, while long-standing outliers are labeled \Oold.  
The integration delay \(\Delay = \Ti - \Ta\) is the time between a document’s first appearance and its first integration into a topic.  
The cutoff \td, which separates these two cases, is set to the empirical 90th percentile of delays among outlier documents that later joined a topic.

A further distinction separates documents that integrate directly into a topic from those that spend time as outliers before integration.
\Tfirst\ and \Tlate, grouped as \TNOcat, join a topic without passing through an outlier stage.
\Tfirst\ joins exactly when the topic forms (\(\Ta=\Ti=\Tt\)), whereas \Tlate\ joins after topic creation (\(\Tt<\Ta=\Ti\)).

The remaining classes concern documents with an explicit outlier phase.  Among them, \TOAcat\ corresponds to the \emph{anticipatory-outlier} pattern (\(\Ta < \Tt \le \Ti\)). These are divided into \TOAfirst, where integration coincides with topic creation (\(\Ti = \Tt\)), and \TOAlate, where it occurs only after the topic is established (\(\Tt < \Ti\)). In contrast, \TODlate\ represents \emph{topic drift}: an outlier that appears after its future topic is created (\(\Tt < \Ta\)), reinforcing an existing topic rather than introducing a new one.

Each document in the corpus is labeled according to this taxonomy. Our analysis focuses on anticipatory cases (\TOAcat) to examine their role as early signals. More broadly, the taxonomy provides a structured view of how news trajectories unfold.

\begin{figure*}[t]
\centering
\small
\resizebox{0.95\textwidth}{!}{%
\begin{forest}
for tree={
  draw, rounded corners, align=center, edge={->},
  s sep=12mm, l sep=8mm, inner xsep=10pt, inner ysep=4pt,
  font=\small
}
[{Joins topic?\\ (\(\exists\, \Ti\))
}
  [{Outlier before \(\Ti\)?\\ (\(\Ta < \Ti\))},
    edge label={node[pos=0.20,left, xshift=-12pt]{\cmark}},%
    label={[yshift=-2pt]below:{\(\Tcat\)}}
    [{\(\Ta < \Tt?\)},
      edge label={node[pos=0.35,left, xshift=-12pt]{\cmark}},
      label={[yshift=-2pt]below:{\(\TOcat\)}}
      [{\(\Ti = \Tt?\)},
        edge label={node[pos=0.35,left, xshift=-10pt]{\cmark}},
        label={[yshift=-2pt]below:{\(\TOAcat\)}}
        [{\(\TOAfirst\)}, seeder, edge={seededge},
          edge label={node[pos=0.35,left, xshift=-10pt]{\cmark}}]
        [{\(\TOAlate\)},  seeder, edge={seededge},
          edge label={node[pos=0.35,right, xshift=10pt]{\xmark}}]
      ]
      [{\(\TODlate\)},
        edge label={node[pos=0.35,right, xshift=10pt]{\xmark}}]
    ]
    [{\(\Ti = \Tt?\)},
      edge label={node[pos=0.35,right, xshift=12pt]{\xmark}},
      label={[yshift=-2pt]below:{\(\TNOcat\)}}
      [{\(\Tfirst\)},
        edge label={node[pos=0.35,left, xshift=-10pt]{\cmark}}]
      [{\(\Tlate\)},
        edge label={node[pos=0.35,right, xshift=10pt]{\xmark}}]
    ]
  ]
  [{\(\Tf - \Ta < \td?\)},
    edge label={node[pos=0.20,right, xshift=12pt]{\xmark}},%
    label={[yshift=-2pt]below:{\(\Ocat\)}}
    [{\(\Orec\)}, edge label={node[pos=0.35,left, xshift=-10pt]{\cmark}}]
    [{\(\Oold\)}, edge label={node[pos=0.35,right, xshift=10pt]{\xmark}}]
  ]
]
\end{forest}
}
\caption{Decision tree mapping conditions on \((\Tt,\Ta,\Ti,\td)\) to the seven cases. \emph{Left} branch splits outlier-integrations \((\TOAfirst,\TOAlate,\TODlate)\) from non-outlier integrations \((\Tfirst,\Tlate)\); \emph{right} branch yields outliers \((\Orec,\Oold)\) persisting until \(\Tf\). \text{\cmark} / \text{\xmark} indicate branch outcomes.}
\label{fig:tree-cases-equations}
\end{figure*}
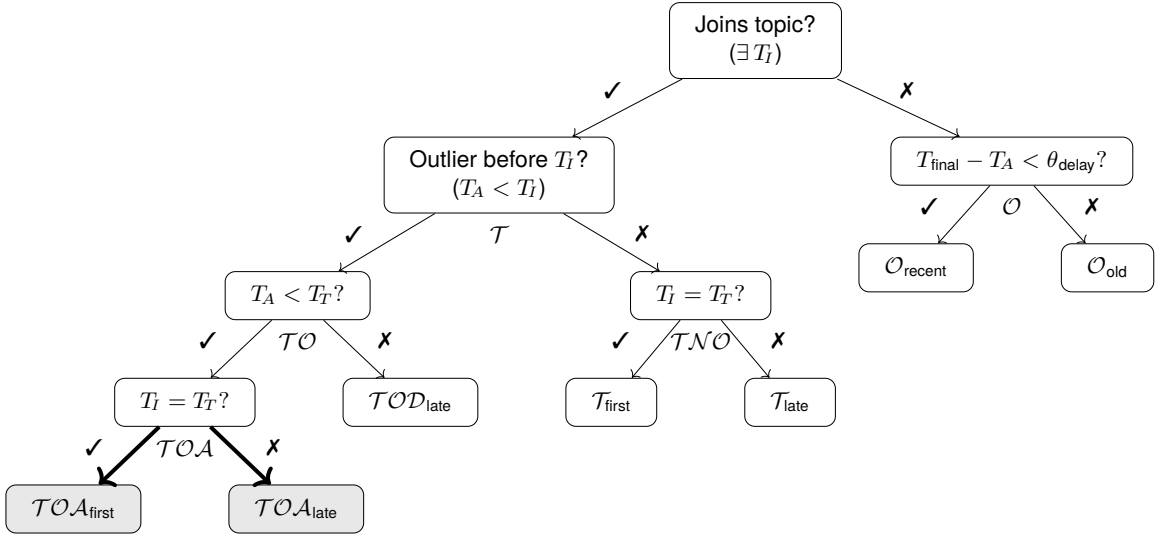

\begin{figure*}[t]%
\centering
\begin{subfigure}{0.32\textwidth}
\centering
\begin{tikzpicture}[x=0.6cm,y=0.7cm,>=stealth,
                    every node/.style={text=black},
                    baseline={(current bounding box.center)}]
  \timelineaxis
  \path[use as bounding box] (-0.1,-1.05) rectangle (5.6,0.95);

  \topicbelt{3.0}{5.5}
  \outlierbelt{1.2}{3.0}
  \draw[gray!70] (3.0,0) -- +(0,\TickUp) node[above=2pt,xshift=-3pt]{\(\Tt\)};
  \fill (1.2,0) circle (1.9pt) node[below=5pt]{\(\Ta\)};
  \fill (3.0,0) circle (2.2pt) node[below=5pt]{\(\Ti\)};
\end{tikzpicture}
\caption{\small \TOAfirst}
\end{subfigure}\hfill
\begin{subfigure}{0.32\textwidth}
\centering
\begin{tikzpicture}[x=0.6cm,y=0.7cm,>=stealth,
                    every node/.style={text=black},
                    baseline={(current bounding box.center)}]
  \timelineaxis
  \path[use as bounding box] (-0.1,-1.05) rectangle (5.6,0.95);

  \topicbelt{2.4}{5.5}
  \outlierbelt{1.0}{4.2}
  \draw[gray!70] (2.4,0) -- +(0,\TickUp) node[above=2pt,xshift=-3pt]{\(\Tt\)};
  \fill (1.0,0) circle (1.9pt) node[below=5pt]{\(\Ta\)};
  \fill (4.2,0) circle (2.2pt) node[below=5pt]{\(\Ti\)};
\end{tikzpicture}
\caption{\small \TOAlate}
\end{subfigure}\hfill
\begin{subfigure}{0.32\textwidth}
\centering
\begin{tikzpicture}[x=0.6cm,y=0.7cm,>=stealth,
                    every node/.style={text=black},
                    baseline={(current bounding box.center)}]
  \timelineaxis
  \path[use as bounding box] (-0.1,-1.05) rectangle (5.6,0.95);

  \topicbelt{1.0}{5.5}
  \outlierbelt{1.8}{4.4}
  \draw[gray!70] (1.0,0) -- +(0,\TickUp) node[above=2pt,xshift=-3pt]{\(\Tt\)};
  \fill (1.8,0) circle (1.9pt) node[below=5pt]{\(\Ta\)};
  \fill (4.4,0) circle (2.2pt) node[below=5pt]{\(\Ti\)};
\end{tikzpicture}
\caption{\small \TODlate}
\end{subfigure}

\vspace{6pt}

\begin{subfigure}{0.32\textwidth}
\centering
\begin{tikzpicture}[x=0.6cm,y=0.7cm,>=stealth,
                    every node/.style={text=black},
                    baseline={(current bounding box.center)}]
  \timelineaxis
  \path[use as bounding box] (-0.1,-1.05) rectangle (5.6,0.95);

  \topicbelt{3.0}{5.5}
  \draw[gray!70] (3.0,0) -- +(0,\TickUp) node[above=2pt,xshift=-3pt]{\(\Tt\)};
  \fill (2.4,0)  node[below=5pt]{\(\Ta\)};
  \fill (3.0,0) circle (2.2pt) node[below=5pt]{\(\Ti\)};
\end{tikzpicture}
\caption{\small \Tfirst}
\end{subfigure}%
\begin{subfigure}{0.32\textwidth}
\centering
\begin{tikzpicture}[x=0.6cm,y=0.7cm,>=stealth,
                    every node/.style={text=black},
                    baseline={(current bounding box.center)}]
  \timelineaxis
  \path[use as bounding box] (-0.1,-1.05) rectangle (5.6,0.95);

  \topicbelt{1.0}{5.5}
  \draw[gray!70] (1.0,0) -- +(0,\TickUp) node[above=2pt,xshift=-3pt]{\(\Tt\)};
  \fill (1.6,0)  node[below=5pt]{\(\Ta\)};
  \fill (2.2,0) circle (2.2pt) node[below=5pt]{\(\Ti\)};
\end{tikzpicture}
\caption{\small \Tlate}
\end{subfigure}\hfill

\vspace{6pt}

\begin{subfigure}{0.32\textwidth}
\centering
\begin{tikzpicture}[x=0.6cm,y=0.7cm,>=stealth,
                    every node/.style={text=black},
                    baseline={(current bounding box.center)}]
  \timelineaxis
  \path[use as bounding box] (-0.1,-1.05) rectangle (5.6,0.95);

  \draw[densely dashed,gray!70] (4.5,-0.5) -- (4.5,0.5)
        node[above=2pt,xshift=-3pt]{\(\td\)};
  \outlierbelt{4.80}{5.5}
  \fill (4.80,0) circle (1.9pt)
        node[below=14pt, anchor=north east, xshift=-2pt, labelbg]{\(\Ta\)};
  \node at (5.5,0) [below=4pt, anchor=north west, xshift=-3pt, labelbg]{\(\Tf\)};
\end{tikzpicture}
\caption{\small \Orec}
\end{subfigure}
\begin{subfigure}{0.32\textwidth}
\centering
\begin{tikzpicture}[x=0.6cm,y=0.7cm,>=stealth,
                    every node/.style={text=black},
                    baseline={(current bounding box.center)}]
  \timelineaxis
  \path[use as bounding box] (-0.1,-1.05) rectangle (5.6,0.95);

  \draw[densely dashed,gray!70] (4.5,-0.5) -- (4.5,0.5)
        node[above=2pt]{\(\td\)};
  \outlierbelt{1.0}{5.5}
  \fill (1.0,0) circle (1.9pt) node[below=5pt]{\(\Ta\)};
  \node at (5.5,0) [below=4pt, anchor=north west, xshift=-3pt, labelbg]{\(\Tf\)};
\end{tikzpicture}
\caption{\small \Oold}
\end{subfigure}

\caption{Overview of the seven taxonomy cases. Panels (a–g) show temporal relations between \Ta, \Tt, \Ti, \td, \Tf. Colored belts indicate topic activity and outlier phases.}

\label{fig:taxonomy-glance-seven}

\end{figure*}
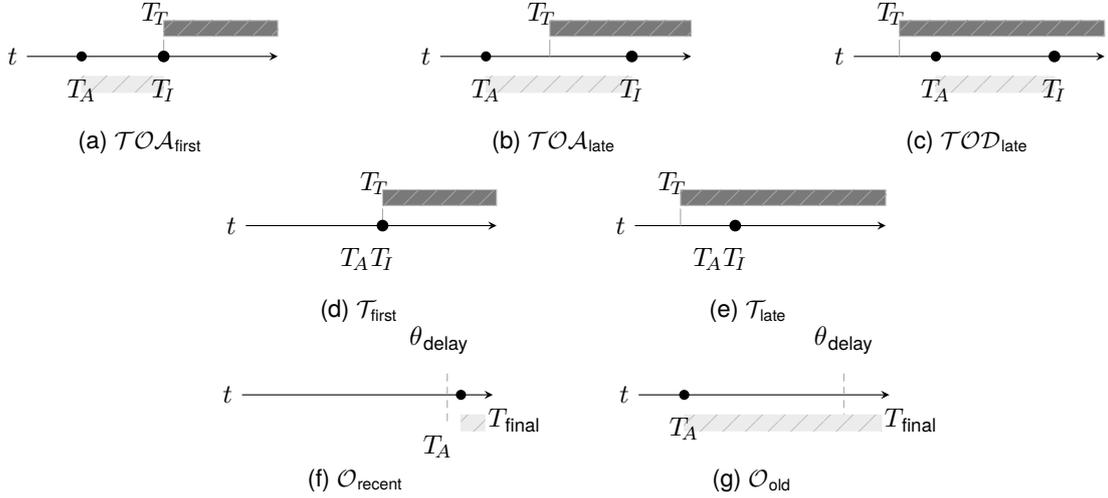

\section{Dataset} 
\label{sec:dataset}
To study the document \emph{trajectories} discussed in the previous section, we required a corpus that is both temporally consistent and topically continuous. Existing news collections often rely on a single source or include irregular sampling, creating gaps in coverage. To address this, we built \HN{}, a French-language corpus on hydrogen energy that provides uninterrupted daily reporting across 81 consecutive days (20~March--8~June~2025).

The dataset captures a period of industrial and policy activity related to hydrogen, including new funding schemes, infrastructure projects, financial developments, and vehicle launches. To ensure source diversity and avoid sparsity, it combines material from official communications, mainstream media, local, and specialized press.

Data were collected daily from two complementary sources to ensure broad and diverse coverage. We retrieved social media posts from \texttt{X (formerly Twitter)}\footnote{\url{https://developer.twitter.com/en/docs/twitter-api}} 
containing the French keyword \emph{hydrogène} (\textit{hydrogen}) and linking to news articles, and articles gathered using the \texttt{Google News}\footnote{\url{https://github.com/ranahaani/GNews}} 
Python library with the same keyword. From \texttt{X}, we collected 1{,}533 posts corresponding to 726 unique articles, with headline and description fields available directly in the API response. \texttt{Google News} yielded an additional 891 unique articles, from which the same fields were extracted. For each item, we recorded metadata such as publication date, URL, and, when available, the country. We applied standard preprocessing, removing boilerplate text (e.g., irrelevant content, repeated templates) and normalizing metadata fields (e.g., date formats). As a result, each document consists of a headline and short description, averaging 280 characters. Finally, we performed cross-source deduplication, prioritizing records with more complete content when overlaps occurred. The final dataset comprises 1{,}616 articles.

\begin{figure}[H]
  \centering
  \includegraphics[width=0.95\linewidth]{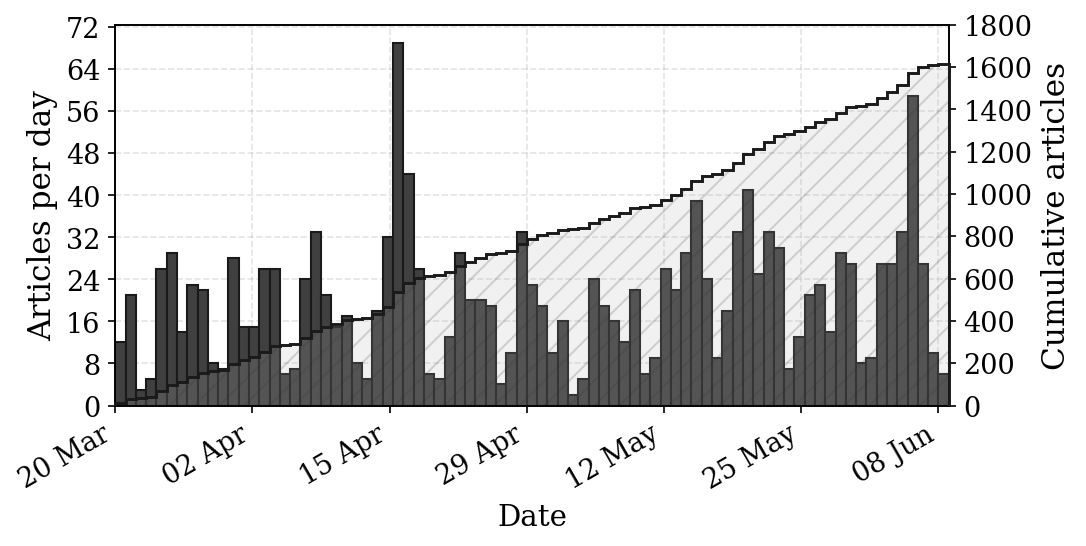}
  \caption{Temporal distribution of \HN{} over the collection period (20~March--8~June~2025). Bars show daily publication counts; the stepped line and hatched area show cumulative totals.}
  \label{fig:dataset_timeline}
\end{figure}

Figure~\ref{fig:dataset_timeline} shows the dataset's temporal distribution. Daily counts reveal coverage peaks and clear regularity with minimal timeline gaps. Cumulative counts confirm uninterrupted coverage.

\section{Modeling Framework}
\label{sec:framework}

To assign each article to a taxonomy case according to whether and when it becomes integrated into a topic, we propose a cumulative clustering framework.
This approach first maps articles to semantic embeddings and then clusters them over successive time windows to capture topic continuity and emergence.

\subsection{Text Representation}
\label{sec:text_representation}

Each news article is represented by embeddings produced by eleven pre-trained language models. For each article, the title and short description are concatenated and encoded as a single text sequence. In the resulting semantic vector space, thematically similar articles are expected to lie close to one another.

To mitigate the \textit{curse of dimensionality} and improve clustering efficiency, we employ Uniform Manifold Approximation and Projection (\texttt{UMAP})~\cite{mcinnes2018umap} to obtain lower-dimensional representations of the embeddings.\footnote{\url{https://umap-learn.readthedocs.io}}
We evaluate output dimensionalities of 2, 3, 5, 10, 20, 30, and 40, keeping the default hyperparameters and fixing the random seed for reproducibility.

We selected eleven embedding models representing diverse architectures, based on their performance on the Massive Text Embedding Benchmark (MTEB)~\cite{muennighoff2023mteb} as of June~2025.
For our French-language corpus, we prioritized multilingual and French-specialized models. The final set includes both open-source models available on Hugging Face and proprietary models.%

\begin{table}[H]
\centering
\scriptsize
\setlength{\tabcolsep}{1.5pt}
\renewcommand{\arraystretch}{0.88}

\begin{tabularx}{\columnwidth}{@{}>{\raggedright\arraybackslash}X
                                >{\centering\arraybackslash}p{1.50cm}
                                >{\centering\arraybackslash}p{1.50cm}@{}}
\toprule
\textbf{Model} & \textbf{Dimensions} & \textbf{Language} \\
\midrule
\href{https://huggingface.co/dangvantuan/sentence-camembert-base}{\texttt{sentence-camembert-base}} & 768  & French \\
\href{https://huggingface.co/OrdalieTech/Solon-embeddings-large-0.1}{\texttt{Solon-embeddings-large-0.1}} & 1024 & French \\
\href{https://huggingface.co/sentence-transformers/paraphrase-multilingual-MiniLM-L12-v2}{\texttt{paraphrase-..-MiniLM-L12-v2}} & 384  & Multilingual \\
\href{https://huggingface.co/sentence-transformers/paraphrase-multilingual-mpnet-base-v2}{\texttt{paraphrase-..-mpnet-base-v2}} & 768  & Multilingual \\
\href{https://huggingface.co/sentence-transformers/LaBSE}{\texttt{LaBSE}} & 768  & Multilingual \\
\href{https://huggingface.co/intfloat/multilingual-e5-large}{\texttt{multilingual-e5-large}} & 1024 & Multilingual \\
\href{https://huggingface.co/Snowflake/snowflake-arctic-embed-l-v2.0}{\texttt{snowflake-arctic-embed-l-v2.0}} & 1024 & Multilingual \\
\href{https://huggingface.co/BAAI/bge-m3}{\texttt{bge-m3}} & 1024 & Multilingual \\
\href{https://platform.openai.com/docs/guides/embeddings}{\texttt{text-embedding-3-small}} & 1536 & Multilingual \\
\href{https://deepmind.google/technologies/gemini}{\texttt{gemini-embedding-001}} & 3072 & Multilingual \\
\href{https://docs.mistral.ai}{\texttt{mistral-embed}} & 1024 & Multilingual \\
\bottomrule
\end{tabularx}

\caption{The eleven embedding models.}
\label{tab:models}
\end{table}

\subsection{Cumulative Clustering}
\label{sec:cumulative_clustering}

Our clustering procedure adapts the \texttt{BERTopic}~\cite{grootendorst2022bertopic} approach with a cumulative design that incrementally incorporates prior documents to model topic evolution. 
We cluster reduced embeddings from the beginning of the corpus up to each day, producing a sequence of expanding windows that captures its evolving topical structure. The study period (20 Mar--8 Jun 2025) yields 81 daily windows. This daily granularity aligns with the pace of the news cycle.

We use two density-based clustering algorithms, \texttt{HDBSCAN}\footnote{\url{https://hdbscan.readthedocs.io}} and \texttt{OPTICS}\footnote{\url{https://scikit-learn.org/stable/modules/generated/sklearn.cluster.OPTICS}}.
Both algorithms were run with their default parameters, and a fixed random seed for reproducibility, each assigning documents a cluster ID or outlier label ($-1$). In \texttt{HDBSCAN}, outlier detection relies on the \texttt{GLOSH} algorithm, which estimates relative local density and designates documents in sparse regions as outliers. In \texttt{OPTICS}, outliers are inferred from the reachability plot when a document’s reachability distance exceeds the cluster-boundary threshold.

Clustering quality is assessed using the silhouette score~\cite{shahapure2020cluster}, which measures intra-cluster cohesion versus inter-cluster separation. Scores above 0.7 indicate strong structure, 0.5--0.7 moderate structure, and below 0.25 weak structure. We report mean and median silhouette scores across all time windows for all combinations of embedding models, clustering algorithms, and \texttt{UMAP} dimensionalities.

\subsection{Topic Alignment}
\label{sec:topic_alignment}

Since the cumulative clustering pipeline produces clusters interpreted as topics independently at each time window, a linking method is required to trace their trajectories. Unlike more flexible approaches such as ANTM~\cite{rahimi2024antm}, which allow many-to-many mappings, we enforce one-to-one correspondences to preserve interpretability. Each previous topic either continues or disappears, and new topics can emerge~\cite{vaca2014time}. Splits and merges are treated as discontinuities.

Formally, each topic cluster is represented by the centroid of its \texttt{UMAP}-reduced embeddings. Let  
\[
T^{(t)} = \{\tau^{(t)}_1, \dots, \tau^{(t)}_{n_t}\}
\]
be the set of topics (i.e., clusters) at time \(t\), with centroid  
\[
\mathbf{c}^{(t)}_i = 
\frac{1}{|\mathcal{D}^{(t)}_i|} 
\sum_{\mathbf{d} \in \mathcal{D}^{(t)}_i} \mathbf{d},
\]
where \(\mathcal{D}^{(t)}_i\) is the set of document embeddings assigned to topic \(\tau^{(t)}_i\), excluding outliers.  

To align topics across consecutive windows, we compute the cosine distance matrix:  
\[
D_{ij} = 1 - 
\frac{\mathbf{c}^{(t-1)}_i \cdot \mathbf{c}^{(t)}_j}
     {\|\mathbf{c}^{(t-1)}_i\| \, \|\mathbf{c}^{(t)}_j\|},
\]
and apply the Hungarian algorithm~\cite{kuhn1955hungarian} to obtain the optimal one-to-one matching.  
A distance threshold \(\theta_{\text{align}}\) determines whether a topic continues over time: if \(D_{ij} \leq \theta_{\text{align}}\), topic \(\tau^{(t)}_j\) is matched to \(\tau^{(t-1)}_i\); otherwise, it is treated as a new topic.  
We vary \(\theta_{\text{align}}\) from $0.20$ to $0.70$ in increments of $0.10$ to assess sensitivity. Lower values yield stricter matching and more splits, whereas higher values allow looser matching and fewer new topics.

\subsection{Case Assignment}
\label{sec:case-assign}

For each configuration, defined by the \texttt{UMAP} dimensionality, clustering algorithm, and \(\ta\), we track the trajectory of every document across daily windows.  
Each of the eleven embedding models assigns one taxonomy case to every document, producing a document–model label matrix.%

\section{Experiments}
\label{sec:results}

\subsection{Topic-based Clustering}
\label{sec:topic-results}

Following the methodology described in Section~\ref{sec:framework}, we performed cumulative clustering over 81 daily windows in the \HN{} corpus. We tested eleven embedding models, seven \texttt{UMAP} dimensions (2--40D), two clustering algorithms (\texttt{HDBSCAN} and \texttt{OPTICS}), and six topic-alignment thresholds (\ta) ranging from $0.20$ to $0.70$.

Clustering quality was evaluated for each configuration using the mean silhouette score across all cumulative time windows. The results in Table~\ref{tab:silhouette_table} indicate moderate to strong cluster separation, with scores ranging from $0.419$ to $0.653$ for \texttt{HDBSCAN} and from $0.551$ to $0.643$ for \texttt{OPTICS}. Across embedding models, both algorithms reach their highest mean and median values at 2D \texttt{UMAP} reduction. \texttt{mistral-embed} yields the highest silhouette scores at all dimensionalities for \texttt{HDBSCAN}, whereas for \texttt{OPTICS} the top score varies across dimensionalities. Overall, the best configuration is \texttt{UMAP} 5D with \texttt{HDBSCAN} on \texttt{mistral-embed}, achieving a silhouette score of $0.653$.

\begin{table*}[t]
\centering
\fontsize{7.2}{7.8}\selectfont
\setlength{\tabcolsep}{3.5pt}
\renewcommand{\arraystretch}{0.96}

\begin{tabular}{lccccccc @{\hspace{6pt}\vrule\hspace{6pt}} ccccccc}
\toprule
\multirow{2}{*}{\textbf{Model}} &
\multicolumn{7}{c}{\textbf{HDBSCAN}} &
\multicolumn{7}{c}{\textbf{OPTICS}} \\
\cmidrule(lr){2-8}\cmidrule(lr){9-15}
& 2D & 3D & 5D & 10D & 20D & 30D & 40D
& 2D & 3D & 5D & 10D & 20D & 30D & 40D \\
\midrule

bge-m3
& 0.621 & 0.611 & 0.589 & 0.618 & 0.583 & 0.602 & 0.615
& 0.634 & 0.615 & 0.610 & 0.592 & \textbf{0.637} & 0.602 & \textbf{0.638} \\
sentence-camembert-base
& 0.582 & 0.579 & 0.568 & 0.562 & 0.574 & 0.559 & 0.561
& 0.612 & 0.565 & 0.585 & 0.551 & 0.600 & 0.571 & 0.578 \\
gemini
& 0.601 & 0.581 & 0.575 & 0.576 & 0.574 & 0.578 & 0.565
& 0.633 & 0.605 & 0.619 & 0.598 & 0.626 & 0.621 & 0.602 \\
multilingual-e5-large
& 0.624 & 0.604 & 0.616 & 0.603 & 0.617 & 0.601 & 0.608
& 0.635 & 0.624 & 0.612 & 0.597 & 0.620 & 0.620 & 0.611 \\
mistral-embed
& \textbf{0.638} & \textbf{0.628} & \textbf{0.653} & \textbf{0.641} & \textbf{0.636} & \textbf{0.636} & \textbf{0.630}
& 0.638 & \textbf{0.643} & \textbf{0.636} & \textbf{0.606} & 0.620 & \textbf{0.628} & 0.617 \\
text-embedding-3-small
& 0.606 & 0.611 & 0.608 & 0.602 & 0.597 & 0.600 & 0.596
& \textbf{0.639} & 0.612 & 0.612 & 0.603 & 0.618 & 0.609 & 0.614 \\
Solon-embeddings-large-0.1
& 0.606 & 0.612 & 0.601 & 0.583 & 0.589 & 0.594 & 0.590
& 0.628 & 0.625 & 0.601 & 0.567 & 0.610 & 0.600 & 0.592 \\
LaBSE
& 0.596 & 0.603 & 0.577 & 0.559 & 0.553 & 0.552 & 0.568
& 0.599 & 0.607 & 0.611 & 0.573 & 0.592 & 0.593 & 0.579 \\
paraphrase..-MiniLM-L12-v2
& 0.519 & 0.496 & 0.565 & 0.551 & 0.547 & 0.483 & 0.419
& 0.603 & 0.571 & 0.566 & 0.545 & 0.589 & 0.565 & 0.564 \\
paraphrase..-mpnet-base-v2
& 0.514 & 0.545 & 0.567 & 0.544 & 0.564 & 0.565 & 0.553
& 0.607 & 0.580 & 0.574 & 0.577 & 0.570 & 0.577 & 0.573 \\
snowflake-arctic..-l-v2.0
& 0.612 & 0.621 & 0.594 & 0.619 & 0.613 & 0.602 & 0.575
& 0.631 & 0.610 & 0.617 & 0.600 & 0.595 & 0.579 & 0.594 \\

\midrule
\textbf{Mean}
& \textbf{0.593} & 0.590 & 0.592 & 0.587 & 0.586 & 0.579 & 0.571
& \textbf{0.624} & 0.605 & 0.604 & 0.583 & 0.607 & 0.597 & 0.597 \\
\textbf{Median}
& \textbf{0.606} & 0.604 & 0.589 & 0.583 & 0.583 & 0.594 & 0.575
& \textbf{0.631} & 0.611 & 0.610 & 0.585 & 0.614 & 0.601 & 0.597 \\
\bottomrule
\end{tabular}

\caption{Average silhouette scores per model and UMAP dimension. Best score per dimension is bolded.}
\label{tab:silhouette_table}
\end{table*}

\begin{figure}[H]
  \centering
  \includegraphics[width=\linewidth]{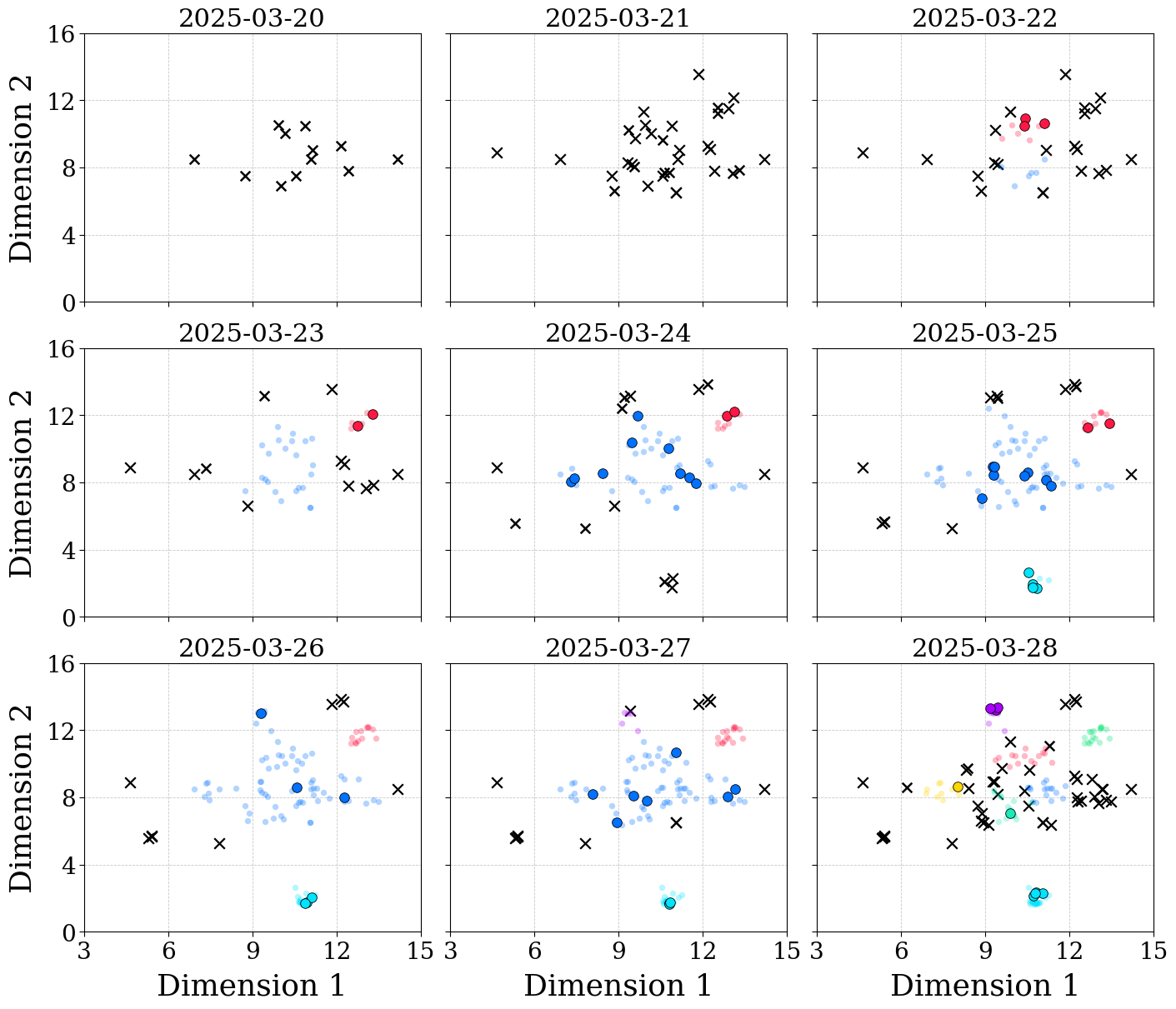}
    \caption{Cumulative clustering over the first 9 time windows with \texttt{mistral-embed} and 2D \texttt{UMAP}. Colors indicate topics; newly assigned documents are larger and more opaque; black $\times$ denote outliers.
    }
  \label{fig:cine-scatter-plot}
\end{figure}

Figure~\ref{fig:cine-scatter-plot} illustrates typical trajectory patterns observed in cumulative clustering. Some articles first appear as outliers and later integrate into a topic, anticipating its emergence (\TOAcat). 
Others appear only after a topic has formed, reflecting drift within established topics (\TODlate). 
A third group integrates directly without an outlier phase (\Tcat).%

\subsection{Topic Integration Delays}
\label{sec:results-delay}

The trajectory analysis reveals when documents transition from outliers to integrated topics. To better understand this temporal dynamic, we quantify how long such transitions take using the notion of \emph{integration delay}, defined as \(\Delay = \Ti - \Ta\) (in days) for an article that first appears as an outlier and later joins a topic. The empirical survival function is given by \(S(t) = P(\Delay > t)\), pooled across models and configurations. Only outlier articles that eventually integrated are included in this analysis (\TOAfirst, \TOAlate, \TODlate).

\begin{figure}[H]
  \centering
  \includegraphics[width=0.95\linewidth]{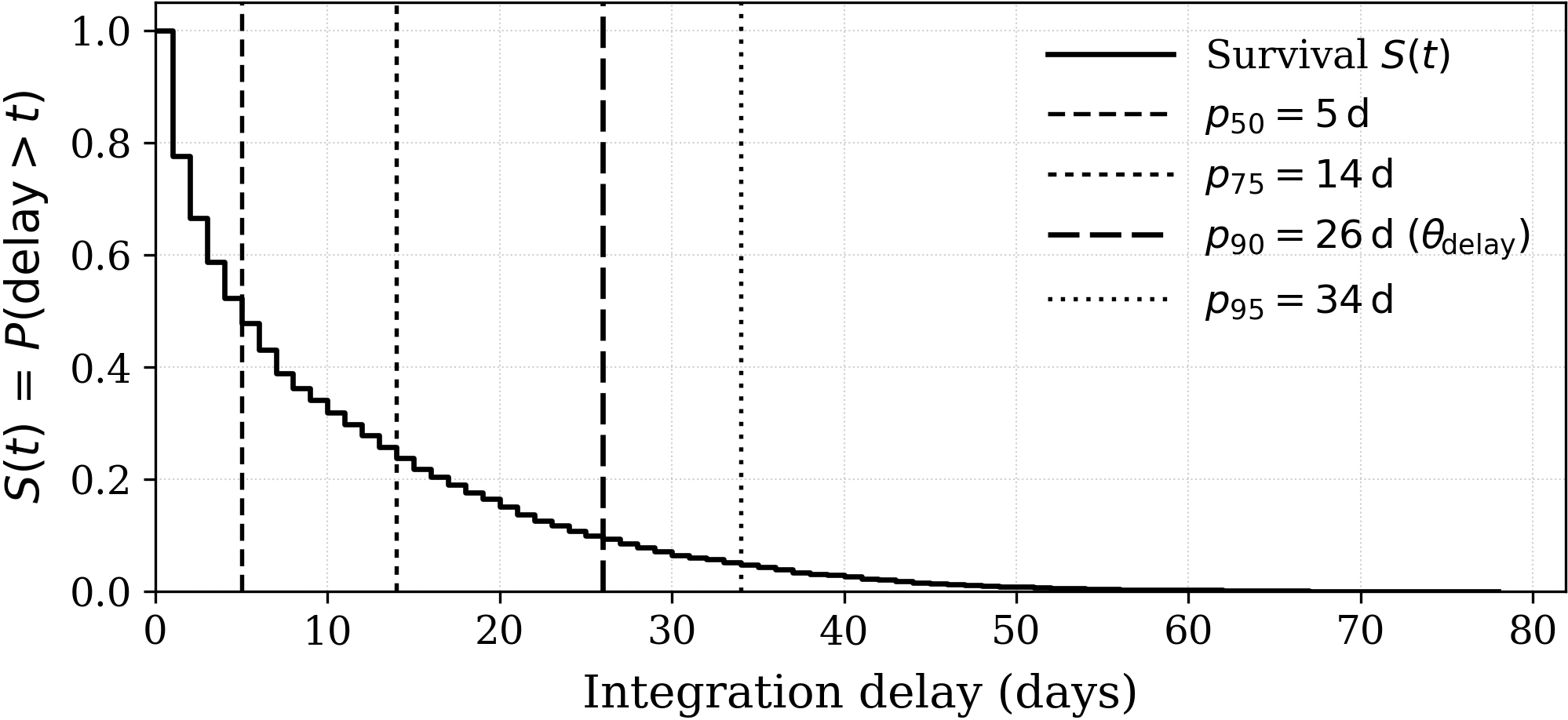}
  \caption{Empirical survival curve \(S(t)=P(\Delay>t)\) for integration delays,
  pooled across models and configurations. Dashed lines mark quantiles; the 90th percentile
  ($p_{90}=26$~days) defines \(\theta_{\text{delay}}\).}
  \label{fig:survival-curve}
\end{figure}

Figure~\ref{fig:survival-curve} shows that the distribution decays rapidly but exhibits a long tail. The median is \(p_{50}=5\) days and the upper quartile \(p_{75}=14\) days; the tail extends to about a month (\(p_{90}=26\) days; \(p_{95}=34\) days).

Based on this empirical profile, we set the delay cutoff to \(\theta_{\text{delay}}=26\) days, at which at least 90\% of integrating documents have joined a topic. Consistent with our taxonomy, we use \(\theta_{\text{delay}}\) to partition non-integrated items (\Ocat) into \Orec\ (recent outliers) and \Oold\ (persistent outliers).

\subsection{Inter-Model Agreement}
\label{sec:results-agreement}

This section examines the sensitivity of \TOAcat\ labels and delay assignments to the choice of embedding model. The \TOAcat\ task is binary (\TOAcat\ vs.\ all other cases), whereas delay is multiclass, representing the temporal offset until integration. Although \TOAcat\ labels vary across embedding models, agreement declines gradually as the number of agreeing models increases, leaving a small high-agreement core of articles. This pattern suggests that inter-model agreement can serve as a proxy for label robustness and help rank anticipatory signals by confidence.

As a first step, we verified that varying the \texttt{UMAP} dimensionality (2--40D) has limited impact on \TOAcat\ assignments. 
As shown in Table~\ref{tab:kappa-dim-summary}, Fleiss'~$\kappa$ for both case and delay remains in the \textit{fair-to-moderate} range (0.34--0.51) across models and is stable across alignment thresholds (variation~$<$~0.01). 
These results suggest that inter-model agreement is largely invariant to \texttt{UMAP} dimensionality, with most variation arising from differences among embedding models. Detailed per-model results are reported in Appendix Table~\ref{tab:interdim_kappa_toa}.

\begin{table}[t]
\centering
\scriptsize
\renewcommand{\arraystretch}{1.1}
\setlength{\tabcolsep}{5pt}
\begin{tabular}{@{}lcccc@{}}
\toprule
Algorithm & Task & Mean Fleiss' $\kappa$ & Range \ta & Model Range\\
\midrule
HDBSCAN & Case & 0.46 & 0.46--0.47 & 0.36--0.51 \\
HDBSCAN & Delay & 0.36 & 0.36--0.37 & 0.20--0.43\\
\hline 
OPTICS & Case & 0.37 & 0.37--0.37 & 0.34--0.41\\
OPTICS & Delay & 0.30 & 0.30--0.31 & 0.23--0.33\\
\bottomrule
\end{tabular}
\caption{Mean Fleiss' $\kappa$ on anticipatory outliers (\TOAcat) across all models and alignment thresholds $\ta \in [0.2,0.7]$. ``Range $\ta$'' shows variability across $\ta$ for fixed models, while ``Model Range'' shows variability across models for fixed $\ta$.}
\label{tab:kappa-dim-summary}
\end{table}

With dimensionality effects shown to be minimal, we focus on variation across embedding models. While Fleiss'~$\kappa$ provides an overall measure of agreement corrected for chance and class imbalance, it does not indicate how consistently individual documents are labeled across embedding models. To capture this document-level consistency, we introduce the \emph{majority agreement} ($MA$), treating the eleven models as independent raters that each assign a label to every article. For each article, we calculate the proportion of models agreeing on the most frequent label, and then average this proportion across all articles:
\[
\qquad \mathrm{MA} = \frac{1}{D}\sum_{i=1}^{D} \mathrm{MA}_i ,\quad \text{with}\ \ \mathrm{MA}_i = \max_k \frac{n_{ik}}{M}\qquad\qquad
\]
where $M=11$ is the number of models, $D$ the number of documents, and $n_{ik}$ the number of models assigning label~$k$, with $k$ denoting a taxonomy case for the \emph{case} task or a delay value for the \emph{delay} task. For the anticipatory-outlier (\emph{case}) task, $k \in \{0,1\}$ represents binary decisions, while for \emph{delay}, $k$ takes multiple discrete values, with $\Delay=0$ for immediate integration and $\Delay=\infty$ for never integrating.

\begin{table*}[t]
\centering
\scriptsize
\renewcommand{\arraystretch}{1.1}
\setlength{\tabcolsep}{5pt}
\begin{minipage}[t]{0.48\textwidth}
\centering
\begin{tabular}{@{}l l c c c@{}}
\toprule
Algorithm & Task & Mean $MA$ & Range & Best UMAP/$\ta$ \\
\midrule
\multirow{2}{*}{HDBSCAN}
 & Case  & 0.94 & 0.94--\textbf{0.95} & 20D / 0.30 \\
 & Delay & 0.74 & 0.72--\textbf{0.76} & 40D$\simeq$20D / 0.30 \\
\midrule
\multirow{2}{*}{OPTICS}
 & Case  & 0.92 & 0.91--\textbf{0.93} & 2D / 0.20 \\
 & Delay & 0.66 & 0.65--\textbf{0.68} & 10D / 0.60 \\
\bottomrule
\end{tabular}
\caption{Mean majority agreement ($MA$) for \TOAcat\ across $\ta \in [0.2,0.7]$ and \texttt{UMAP} dimensions (2--40D). 
The range shows min--max values, and ``Best UMAP/$\ta$'' indicates where the maximum occurs.}

\label{tab:best-MA}
\end{minipage}\hfill
\begin{minipage}[t]{0.48\textwidth}
\centering
\begin{tabular}{@{}l l c c c@{}}
\toprule
Algorithm & Task & Mean Fleiss' $\kappa$ & Range & Best UMAP/$\ta$ \\
\midrule
\multirow{2}{*}{HDBSCAN}
 & Case  & 0.30 & 0.24--\textbf{0.36} & 30D / 0.30 \\
 & Delay & 0.18 & 0.13--\textbf{0.20} & 20D / 0.40 \\
\midrule
\multirow{2}{*}{OPTICS}
 & Case  & 0.20 & 0.13--\textbf{0.26} & 10D / 0.40 \\
 & Delay & 0.16 & 0.09--\textbf{0.24} & 10D / 0.30 \\
\bottomrule
\end{tabular}
\caption{Mean Fleiss' $\kappa$ for \TOAcat\ across $\ta \in [0.2,0.7]$ and \texttt{UMAP} dimensions (2--40D). 
The range shows min--max values, and ``Best UMAP/$\ta$'' indicates where the maximum occurs.}
\label{tab:best-kappa}
\end{minipage}
\end{table*}

Tables~\ref{tab:best-MA} and~\ref{tab:best-kappa} summarize the highest majority agreement ($MA$) across all \texttt{UMAP} dimensionalities (2–40D) and alignment thresholds ($\ta$), along with the corresponding ranges, grouped by clustering algorithm. \texttt{HDBSCAN} systematically outperforms \texttt{OPTICS} on both $MA$ and Fleiss' $\kappa$ for the case and delay tasks. The top configuration is \texttt{HDBSCAN} with \texttt{UMAP}~20D and $\ta=0.30$.

\begin{figure}[H]
  \centering
  \includegraphics[width=0.95\linewidth]{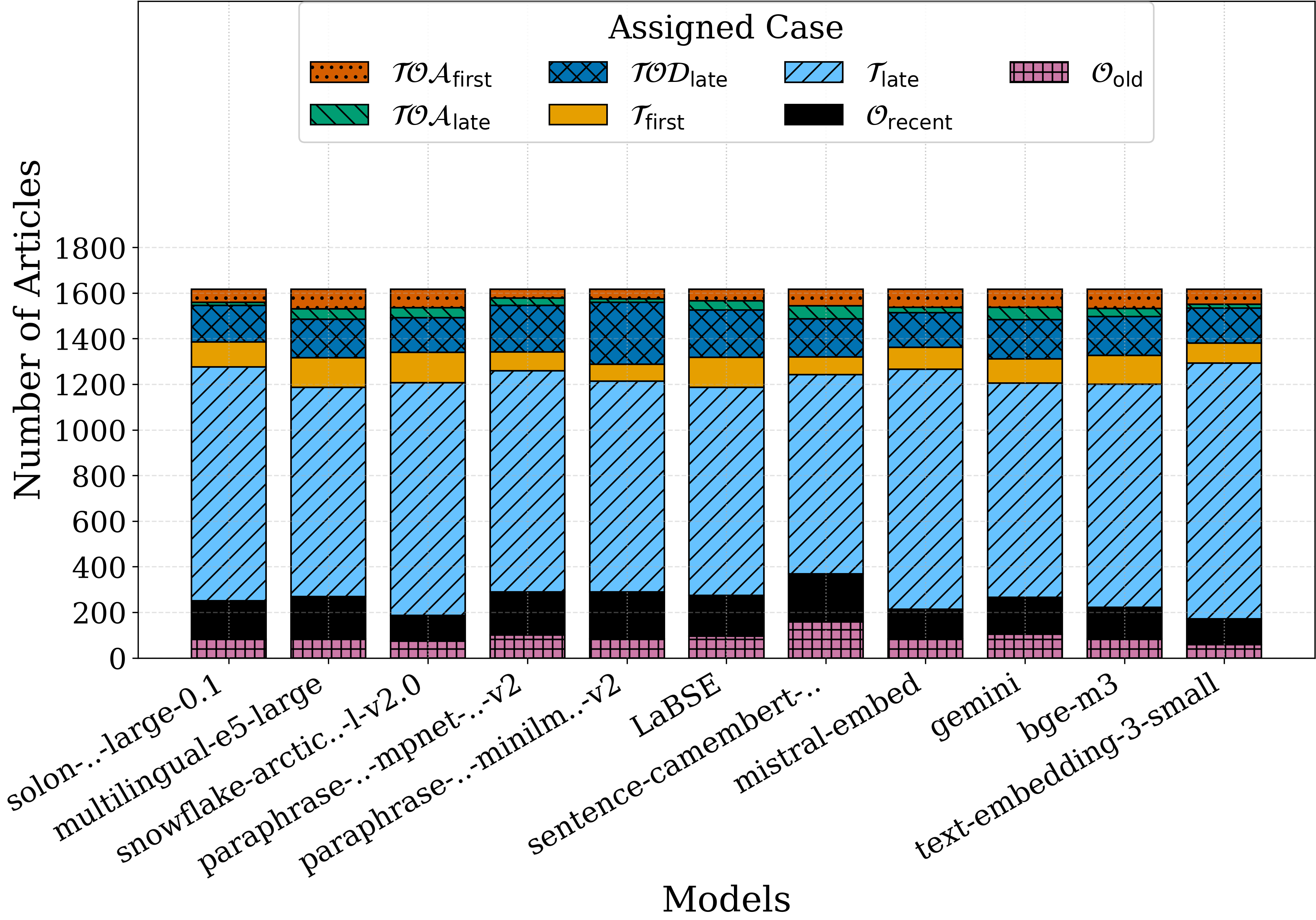}
  \caption{Distribution of taxonomy cases per model for \texttt{HDBSCAN}, \texttt{UMAP} 20D, and $\ta=0.30$.}
  \label{fig:case-distribution-stacked}
\end{figure}

Under this configuration, Figure~\ref{fig:case-distribution-stacked} shows that most articles are assigned to established topics (\Tlate), followed by drift outliers (\TODlate), long-standing or recent non-integrated outliers (\Oold, \Orec), and at-creation integrations (\Tfirst). Anticipatory outliers (\TOAfirst, \TOAlate) are also observed across all eleven models, with a consistent distribution.

\begin{figure}[H]
  \centering
  \includegraphics[width=0.95\linewidth]{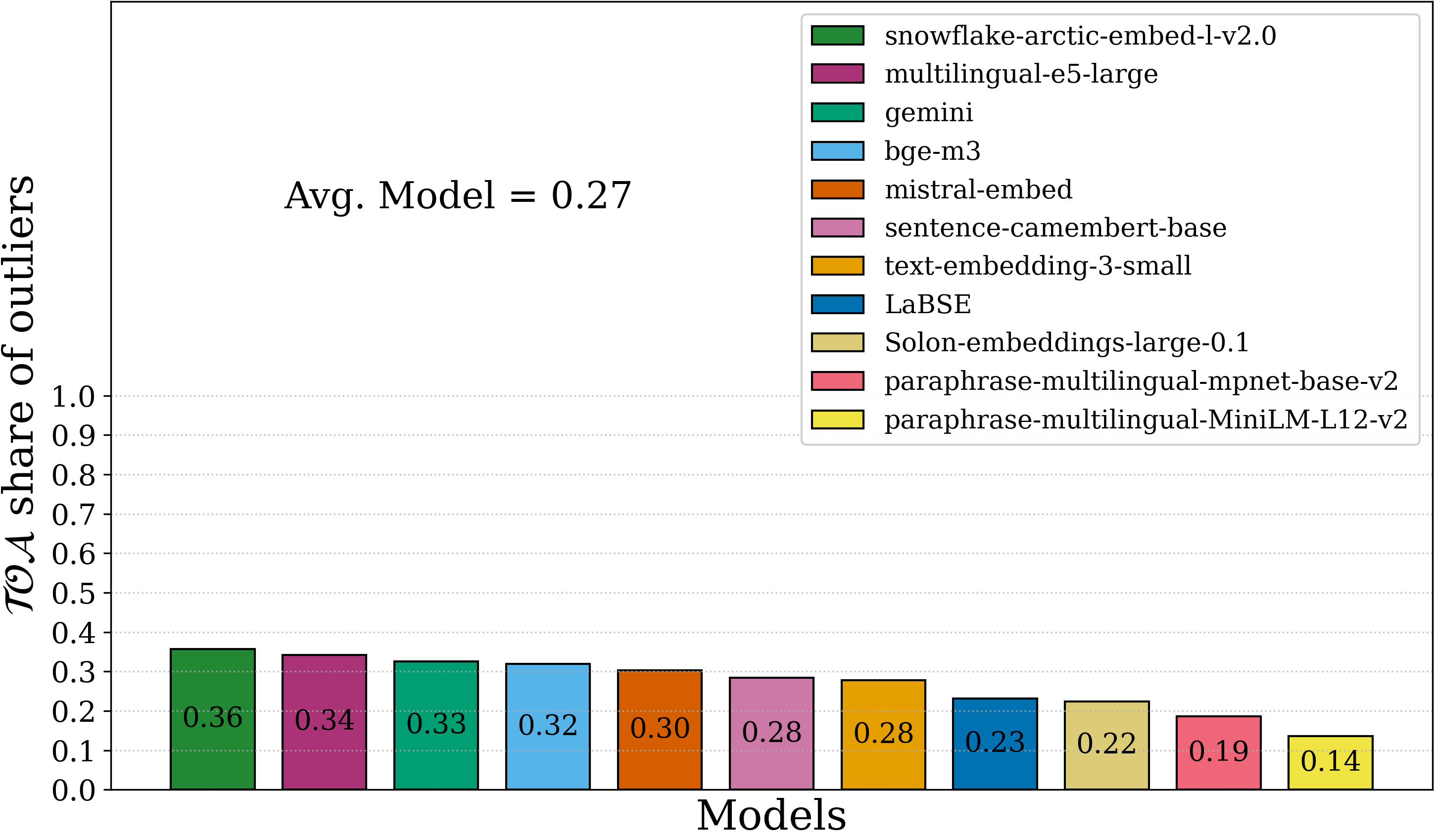}
  \caption{Per-model share of anticipatory outliers (\TOAcat) among all outlier cases (\TOcat, \Ocat)}
  \label{fig:freq-became-seed}
\end{figure}

Figure~\ref{fig:freq-became-seed} shows that, on average, $27\%$ of outliers are anticipatory outliers (\TOAcat).
The highest \TOAcat\ proportions are observed for \texttt{snowflake} ($36\%$), \texttt{e5-large} ($34\%$), \texttt{gemini} ($33\%$), and \texttt{bge-m3} ($32\%$).
In contrast, \texttt{minilm-l12} and \texttt{mpnet} show the smallest shares, $14\%$ and $19\%$, respectively.
For fixed $\ta=0.30$, the average \TOAcat\ proportion is stable across \texttt{UMAP} dimensionalities, ranging from $0.27$ to $0.32$.

\begin{figure}[H]
  \centering
  \begin{subfigure}{\linewidth}
    \includegraphics[width=0.95\linewidth]{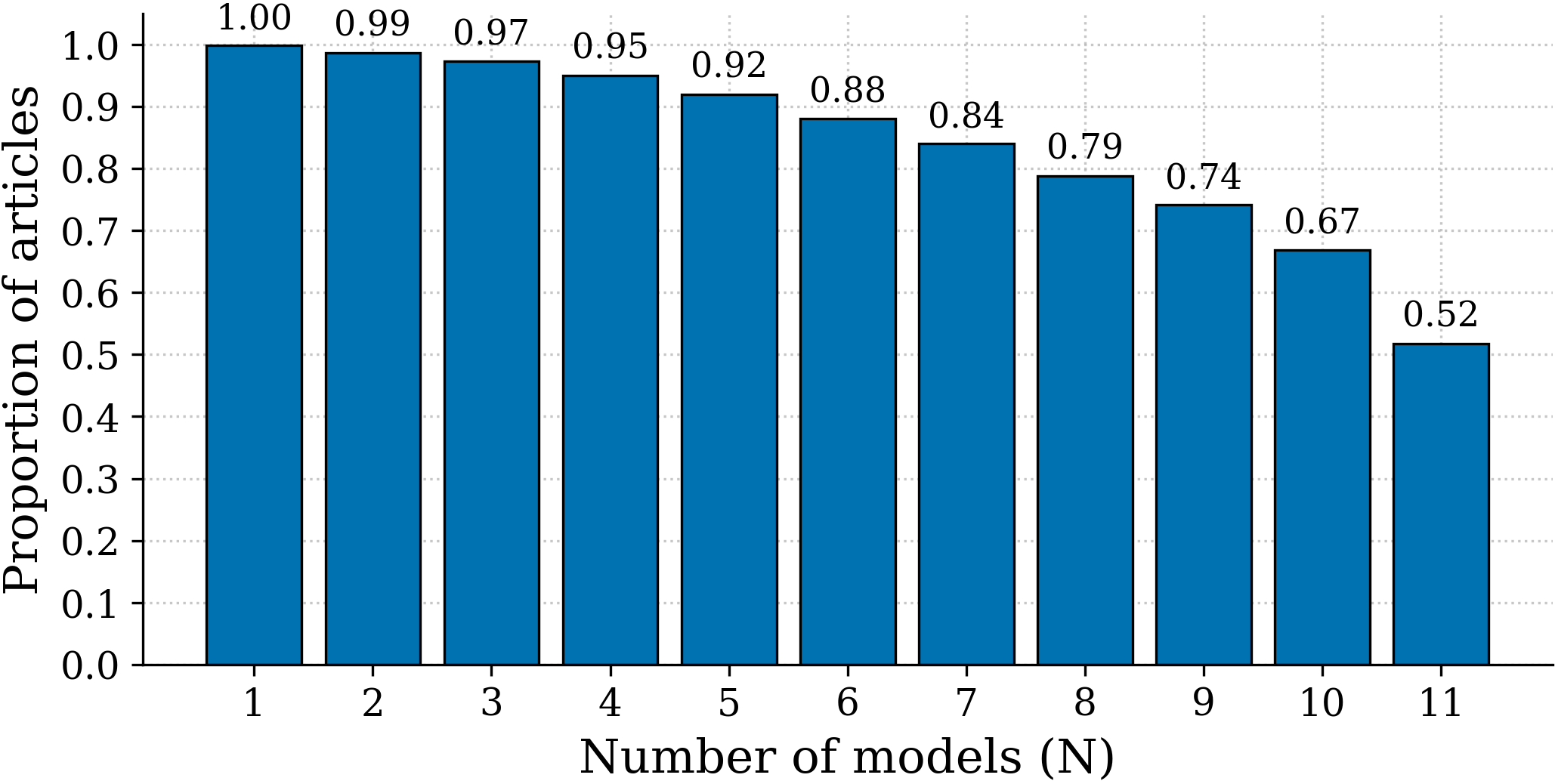}
    \caption{Integrated to Topic (\Tcat).}
    \label{fig:agreement-integration}
  \end{subfigure}

  \vspace{0.05em}

  \begin{subfigure}{\linewidth}
    \includegraphics[width=0.95\linewidth]{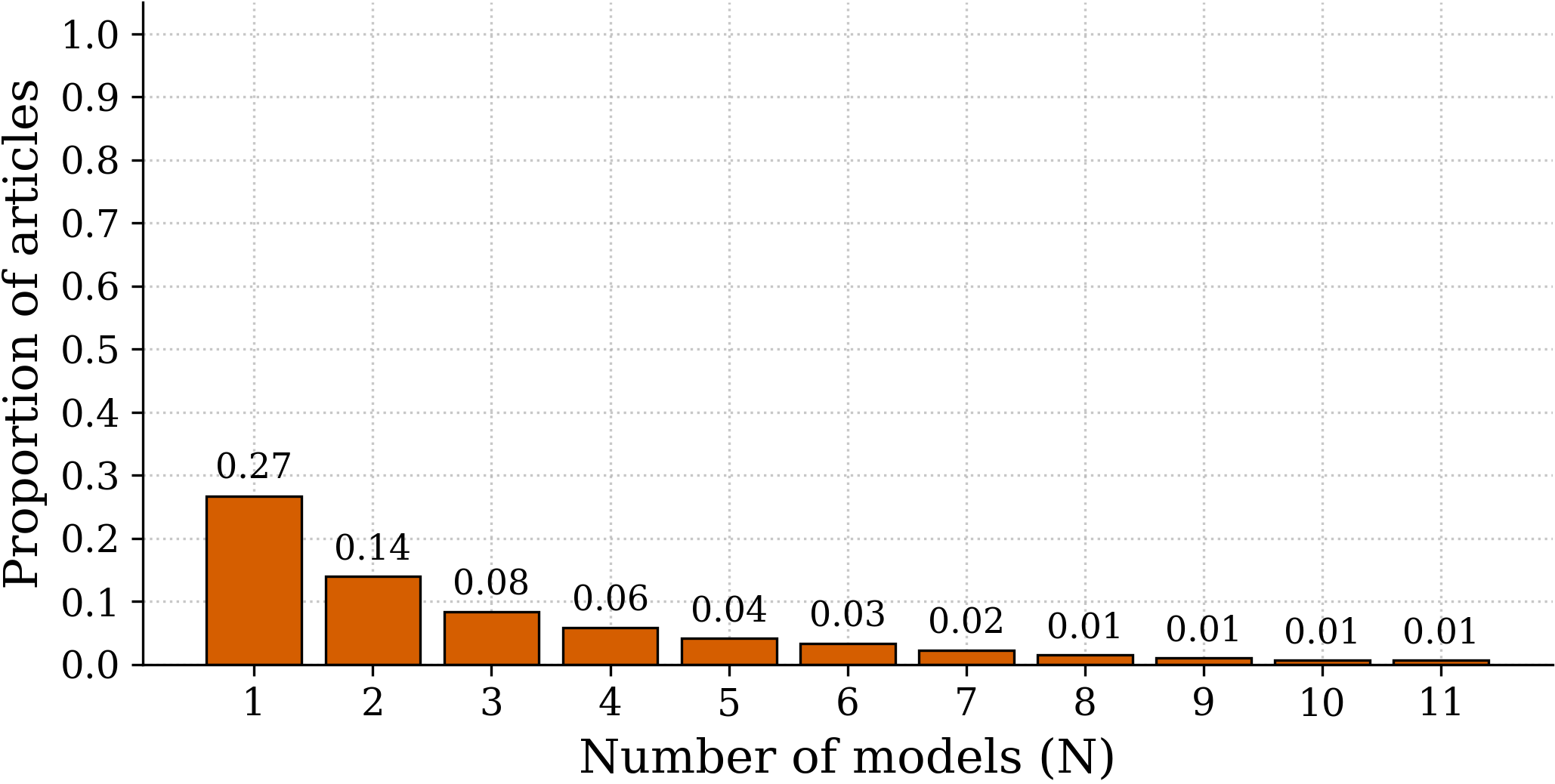}
    \caption{Anticipatory Outliers (\TOAcat).}
    \label{fig:agreement-seeder}
  \end{subfigure}

  \caption{Proportion of articles for which at least $N$ models agree on (a) \Tcat\ and (b) \TOAcat\ cases.}
  \label{fig:agreement-consensus}
\end{figure}

Figures~\ref{fig:agreement-integration} and~\ref{fig:agreement-seeder} show cumulative agreement by the number of agreeing models~(\(N\)). \Tcat\ assignments are highly consistent: at least \(N{=}4\) models agree for \(95\%\) of articles, \(N{=}7\) for \(84\%\), and all eleven for \(52\%\). In contrast, \TOAcat\ unanimity is much lower: at least one model flags \(27\%\) of articles, three models flag \(8\%\), six flag \(3\%\), and all eleven flag about \(1\%\). Although \TOAcat\ agreement is far lower than for \Tcat, it remains a practical measure of robustness. Using thresholds such as \(N\!\ge\!6\), or unanimity, isolates a small, high-confidence core of anticipatory outliers.

\section{Case Studies}
\label{sec:case-studies}

To illustrate the taxonomy's interpretability, we retrospectively analyze three \HN\ topics that include \emph{anticipatory outlier} (\TOAcat) documents.
For each topic, we report majority agreement ($MA$) for both the taxonomy label and the delay under the best-performing configuration reported in Section~\ref{sec:results-agreement} (\texttt{HDBSCAN}, \texttt{UMAP}~20D, $\ta{=}0.3$).
For anticipatory outliers, anticipation is defined as the time difference \(\Tt - \Ta\) (in days).

\paragraph{\textbf{Hyundai NEXO (2025) release.}}
This topic captures the launch of Hyundai's second-generation NEXO hydrogen vehicle, officially revealed on 3~April~2025. The topic forms at \(\Tt=\) 3~April~2025. The earliest signal appears on 21~March at \emph{NewAutoPost} (newautopost.co.kr/fr/), reporting prototype sightings {\small (FR: \emph{``Le véhicule d'essai du NEXO de deuxième génération… repéré.''}, EN: \emph{``Second-generation NEXO test vehicle spotted''})}, anticipating the topic by 13 days and labeled \TOAfirst\ by a majority of models (\(\Ta<\Tt=\Ti\); \(MA_{\text{case}}=11/11\); \(MA_{\text{delay}}=9/11\)). On 2~April, the Canadian outlet \emph{RPM} (rpmweb.ca) published prelaunch specifications {\small (FR: \emph{``NEXO à l'hydrogène est plus puissante, plus spacieuse et propose une autonomie d'environ 700 kilomètres''}, EN: \emph{``The hydrogen NEXO is more powerful, more spacious, and offers about 700 km of range''})}, anticipating the topic formation by 1 day and labeled \TOAfirst\ (\(\Ta<\Tt=\Ti\); \(MA_{\text{case}}=8/11,\; MA_{\text{delay}}=11/11\)). On the release day (3~April), coinciding with topic creation (\Tt) for the majority of models, multiple French media such as \emph{H2-Mobile} {\small (h2-mobile.fr; FR: \emph{``Les spécifications clés sont confirmées ce matin.''}, EN: \emph{``Key specifications are confirmed this morning''})} and the official Hyundai press release {\small (www.hyundai.news; FR: \emph{``Hyundai dévoile NEXO Nouvelle Génération''}, EN: \emph{``Hyundai unveils new-generation NEXO''})} form and stabilize the cluster, typically as \Tfirst\ (\(\Ta=\Tt=\Ti\); \(MA_{\text{case}}=5/11,\; MA_{\text{delay}}=9/11\)). Subsequent coverage following topic formation, for example by \emph{Rouler Electrique} {\small (www.rouler-electrique.fr; FR: \emph{``Hyundai mise encore sur l'hydrogène malgré les doutes scientifiques''}, EN: \emph{``Hyundai is still betting on hydrogen despite scientific doubts''})}, reinforces the topic on 16~April and is predominantly labeled \Tlate\ (\(\Tt<\Ta=\Ti\); \(MA_{\text{case}}=10/11,\; MA_{\text{delay}}=10/11\)).

\paragraph{\textbf{Safra's Financial Crisis.}}
A second case concerns Safra, a French manufacturer of hydrogen buses, which faced severe financial difficulties before its acquisition by the Chinese group Wanrun. The topic forms at \(\Tt=\) 28~April~2025. Early signals in local and national media anticipated the crisis: on 8~April, \emph{France Bleu} {\small (francebleu.fr; FR: \emph{``170 emplois menacés à Safra Albi''}, EN: \emph{``170 jobs threatened at Safra Albi''})} is labeled \TOAfirst\ (\(\Ta<\Tt=\Ti\)), anticipating by 20 days with fair agreement (\(MA_{\text{case}}=7/11,\; MA_{\text{delay}}=5/11\)). On 9~April, \emph{France~3 Occitanie} {\small (france3-regions.francetvinfo.fr; FR: \emph{``avenir incertain des 174 salariés de Safra''}, EN: \emph{``uncertain future for Safra's 174 employees''})} reports an uncertain financial future for Safra and is labeled \TOAfirst\ (\(\Ta<\Tt=\Ti\)) with 19 days of anticipation and fair delay agreement (\(MA_{\text{case}}=8/11,\; MA_{\text{delay}}=5/11\)). On 16~April, \emph{La Tribune} {\small (latribune.fr; FR: \emph{``l'unique constructeur français de bus à hydrogène au bord de la faillite''}, EN: \emph{``the only French hydrogen-bus manufacturer on the brink of bankruptcy''})} accentuates the importance of the risk and is labeled \TOAlate\ (\(\Ta<\Tt<\Ti\)), with 12 days of anticipation (\(MA_{\text{case}}=7/11,\; MA_{\text{delay}}=5/11\)). On the day of topic formation (28~April), \emph{HydrogenToday} {\small (FR: \emph{``Bus à hydrogène : quel avenir pour Safra ?''}, EN: \emph{``Hydrogen buses: what future for Safra?''})} is classified \Tfirst\ (\(\Ta=\Tt=\Ti\)), with low to moderate agreement (\(MA_{\text{case}}=4/11,\; MA_{\text{delay}}=6/11\)), marking the topic's formation. Later coverage, including the 29~April takeover proposal in \emph{La Dépêche} {\small (ladepeche.fr; FR: \emph{``Un groupe asiatique propose de racheter Safra''}, EN: \emph{``An Asian group proposes to buy Safra''})} and the mid-May confirmation of Wanrun's acquisition by \emph{H2-Mobile} {\small (h2-mobile.fr; FR: \emph{``Safra passe sous pavillon chinois''}, EN: \emph{``Safra comes under Chinese ownership''})} is labeled \Tlate\ (\(\Tt<\Ta=\Ti\)), consolidating the topic with high agreement (e.g., \(MA_{\text{case}}\ge 8/11,\; MA_{\text{delay}}=11/11\)).

\paragraph{\textbf{Vallourec DELPHY launch.}}
This topic covers the certification and commercialization of Vallourec's vertical hydrogen storage system \textit{DELPHY}, with coverage concentrated around 5~June~2025.
For most models, topic creation occurs at $\Tt=\text{5~June~2025}$.
Two early articles anticipate the topic by multiple weeks: an interview in \emph{La Tribune Dimanche} on $\Ta=\text{18~May~2025}$ is labeled \TOAfirst\ by a majority of models ($MA_{\text{case}}=8/11$; $MA_{\text{delay}}=9/11$), anticipating the topic by 18 days and satisfying \(\Ta<\Tt=\Ti\).
Similarly, \emph{La Voix du Nord} publishes on $\Ta=\text{21~May~2025}$ and is labeled \TOAfirst\ with higher agreement ($MA_{\text{case}}=9/11$; $MA_{\text{delay}}=10/11$), anticipating the topic by 15 days and again matching \(\Ta<\Tt=\Ti\).
On the announcement day ($\Ta=\Tt=\text{5~June~2025}$), multiple press and market outlets (e.g., \emph{Fortuneo (AOF)}, \emph{Boursorama}, \emph{Investir}, \emph{Les Échos}, \emph{Connaissance des Énergies}, \emph{GlobeNewswire}) form and stabilize the cluster, typically labeled \Tfirst\ ($\Ta=\Tt=\Ti$) with near-unanimous agreement ($MA_{\text{case}}\approx 11/11$; $MA_{\text{delay}}=11/11$).
In this topic, we did not observe \TODlate\ items; integrated documents are either anticipatory outliers (\TOAcat), immediate integrations (\Tfirst), or later integrations (\Tlate).

\smallskip
Across these cases, \TOAfirst{} documents appear strictly before \(\Tt\) and integrate when the topic forms (\(\Ta<\Tt=\Ti\)), providing early evidence that precedes later topic consolidation.
Subsequent \Tfirst{} and \Tlate{} articles stabilize these developments into coherent topics, illustrating how the taxonomy distinguishes anticipatory signals from later mainstream uptake.

\section{Conclusion}
\label{sec:conclusion}

We introduced a taxonomy of news trajectories based on three events: document appearance, topic creation, and first integration, and applied it to the \HN{} corpus using a cumulative clustering framework. This taxonomy formalizes \emph{anticipatory outliers}: documents that predate topic formation yet later integrate, in contrast to those that reinforce existing topics or persist as noise. Across eleven embedding models and parameter settings, agreement was fair to moderate for the binary classification task, with only 1\% of documents unanimously labeled as anticipatory outliers. The best configuration achieved 0.95 majority agreement and a Fleiss' kappa of 0.33.

The agreement analysis showed that combining multiple models helps mitigate label instability across individual models. A small, high-agreement core of articles suggests that inter-model agreement may serve as a useful proxy for label robustness. The case studies further suggest that anticipatory outliers may provide early signals, often appearing days or weeks before a topic forms.

Beyond retrospective analysis, this work provides a basis for prospective early-warning analysis. Future work will extend the approach to predictive modeling of anticipatory outliers as weak signals of emerging topics. We will also examine the drivers of these transitions and the dynamics of other taxonomy cases. Finally, we will assess whether the framework scales to larger corpora and captures broader thematic patterns beyond specific events.

\section*{Limitations}

Our analysis is limited to French-language news. This limits the generalizability of the findings beyond this setting. We use daily aggregation to match the news cycle; this granularity may miss slower and longer-term topic shifts. Future work should test the framework on other languages, domains, larger corpora, and alternative time scales.

\section*{Ethical Considerations}
This study was conducted in line with open-science principles of transparency and reproducibility. When possible, we favor open-source and smaller models in order to reduce computational and environmental costs. The analysis uses short quoted fragments only when necessary for commentary and interpretation, in accordance with the French \emph{exception de courte citation}.

\section*{Data and Code Availability}
To support reproducibility, we make the scripts for corpus construction and the experimental setup available in a dedicated GitHub repository: \url{https://github.com/evangeliazve/lrec_from_noise_to_signal}. The dataset and experimental results can be shared for research purposes on request.

\section*{Acknowledgments}
We thank the three anonymous reviewers for their insightful comments and suggestions. EZ gratefully thanks Infopro Digital for supporting her PhD at LIP6 by allocating part of her work time to it.

\section*{Author Contributions}
EZ, GB, JGG conceived the research problem. EZ designed and conducted the experiments. GB and JGG supervised the research and provided guidance. EZ wrote the paper, and all authors reviewed it and revised it.%

\section*{Bibliographical References}
\label{sec:reference}

\bibliographystyle{lrec2026-natbib}
\bibliography{lrec2026-example}

\appendix
\section{Appendix}
\label{sec:appendix}

\subsection{Case Frequencies}
\label{sec:appendix-frequencies}

Table~\ref{tab:appendix-anticipatory-outliers-combined} reports the share of anticipatory outliers (\TOAcat) among all outlier documents (\TOcat\ $\cup$ \Ocat) across \texttt{UMAP} dimensionalities at $\ta=0.30$. 

\begin{table*}[!ht]
\centering
\scriptsize
\setlength{\tabcolsep}{3.2pt}
\renewcommand{\arraystretch}{1.05}

\begin{adjustbox}{max width=\textwidth}
\begin{tabular}{p{4.6cm} ccccccc @{\hspace{1.2em}} ccccccc}
\toprule
\multirow{2}{*}{\textbf{Model}} &
\multicolumn{7}{c}{\textbf{HDBSCAN}} &
\multicolumn{7}{c}{\textbf{OPTICS}} \\
\cmidrule(lr){2-8}\cmidrule(lr){9-15}
& \textbf{2D} & \textbf{3D} & \textbf{5D} & \textbf{10D} & \textbf{20D} & \textbf{30D} & \textbf{40D}
& \textbf{2D} & \textbf{3D} & \textbf{5D} & \textbf{10D} & \textbf{20D} & \textbf{30D} & \textbf{40D} \\
\midrule
\rowcolor{gray!10}\textbf{Avg.\ Model}
& 0.29 & 0.30 & 0.32 & 0.31 & 0.27 & 0.29 & 0.29
& 0.27 & 0.33 & 0.30 & 0.30 & 0.31 & 0.30 & 0.29 \\
\midrule
bge-m3                                & 0.25 & 0.41 & 0.33 & 0.33 & 0.32 & 0.33 & 0.34 & 0.27 & 0.29 & 0.33 & 0.32 & 0.36 & 0.31 & 0.27 \\
gemini                                & 0.36 & 0.35 & 0.34 & 0.37 & 0.33 & 0.31 & 0.31 & 0.25 & 0.37 & 0.28 & 0.31 & 0.31 & 0.29 & 0.34 \\
LaBSE                                 & 0.32 & 0.30 & 0.37 & 0.31 & 0.23 & 0.24 & 0.30 & 0.31 & 0.29 & 0.27 & 0.24 & 0.30 & 0.26 & 0.30 \\
mistral-embed                         & 0.25 & 0.33 & 0.30 & 0.30 & 0.31 & 0.32 & 0.30 & 0.28 & 0.33 & 0.36 & 0.34 & 0.37 & 0.35 & 0.33 \\
multilingual-e5-large                 & 0.29 & 0.27 & 0.42 & 0.39 & 0.34 & 0.40 & 0.40 & 0.30 & 0.37 & 0.31 & 0.35 & 0.32 & 0.35 & 0.29 \\
paraphrase-multilingual-MiniLM-L12-v2 & 0.39 & 0.38 & 0.27 & 0.11 & 0.14 & 0.17 & 0.22 & 0.29 & 0.30 & 0.27 & 0.24 & 0.23 & 0.25 & 0.23 \\
paraphrase-multilingual-mpnet-base-v2 & 0.25 & 0.13 & 0.18 & 0.46 & 0.19 & 0.22 & 0.19 & 0.25 & 0.29 & 0.29 & 0.24 & 0.25 & 0.28 & 0.24 \\
sentence-camembert-base               & 0.28 & 0.30 & 0.31 & 0.34 & 0.28 & 0.29 & 0.36 & 0.24 & 0.27 & 0.20 & 0.22 & 0.27 & 0.26 & 0.25 \\
snowflake-arctic-embed-l-v2.0         & 0.36 & 0.31 & 0.43 & 0.28 & 0.36 & 0.35 & 0.28 & 0.28 & 0.44 & 0.34 & 0.42 & 0.35 & 0.35 & 0.26 \\
Solon-embeddings-large-0.1            & 0.23 & 0.27 & 0.28 & 0.24 & 0.22 & 0.29 & 0.25 & 0.25 & 0.31 & 0.26 & 0.27 & 0.32 & 0.29 & 0.30 \\
text-embedding-3-small                & 0.26 & 0.28 & 0.28 & 0.26 & 0.27 & 0.29 & 0.29 & 0.30 & 0.33 & 0.39 & 0.33 & 0.36 & 0.35 & 0.33 \\
\bottomrule
\end{tabular}
\end{adjustbox}
\caption{Ratio of \TOAcat\ among outlier documents across \texttt{UMAP} dimensionalities for \texttt{HDBSCAN} and \texttt{OPTICS}, at $\ta=0.30$.}
\label{tab:appendix-anticipatory-outliers-combined}
\end{table*}

\subsection{Agreement Metrics}
\label{sec:appendix-agreement}

Table~\ref{tab:interdim_kappa_toa} reports Fleiss'~$\kappa$ inter-dimensionality agreement by embedding model and $\theta_{\mathrm{align}}$.%

\begin{table*}[!ht]
\centering
\scriptsize
\setlength{\tabcolsep}{3pt}
\renewcommand{\arraystretch}{1.08}

\begin{tabular}{p{4.8cm}
@{\hspace{0.8em}}c@{\hspace{0.9em}}c
@{\hspace{1.1em}}c@{\hspace{0.9em}}c
@{\hspace{1.1em}}c@{\hspace{0.9em}}c
@{\hspace{1.1em}}c@{\hspace{0.9em}}c
@{\hspace{1.1em}}c@{\hspace{0.9em}}c
@{\hspace{1.1em}}c@{\hspace{0.9em}}c
}
\toprule
\multirow{2}{*}{\textbf{Model}} &
\multicolumn{2}{c}{\textbf{$\ta=0.2$}} &
\multicolumn{2}{c}{\textbf{$\ta=0.3$}} &
\multicolumn{2}{c}{\textbf{$\ta=0.4$}} &
\multicolumn{2}{c}{\textbf{$\ta=0.5$}} &
\multicolumn{2}{c}{\textbf{$\ta=0.6$}} &
\multicolumn{2}{c}{\textbf{$\ta=0.7$}} \\
\cmidrule(lr){2-3}\cmidrule(lr){4-5}\cmidrule(lr){6-7}\cmidrule(lr){8-9}\cmidrule(lr){10-11}\cmidrule(lr){12-13}
& $\kappa$ & $\kappa_{\textit{delay}}$
& $\kappa$ & $\kappa_{\textit{delay}}$
& $\kappa$ & $\kappa_{\textit{delay}}$
& $\kappa$ & $\kappa_{\textit{delay}}$
& $\kappa$ & $\kappa_{\textit{delay}}$
& $\kappa$ & $\kappa_{\textit{delay}}$ \\
\midrule

\multicolumn{13}{c}{\textbf{HDBSCAN}}\\
\midrule
bge-m3                      & \textbf{0.51} & 0.38 & \textbf{0.51} & 0.38 & \textbf{0.51} & 0.38 & \textbf{0.51} & 0.38 & \textbf{0.51} & 0.38 & 0.51 & 0.38 \\
sentence-camembert-base     & 0.47 & 0.32 & 0.47 & 0.32 & 0.47 & 0.32 & 0.47 & 0.32 & 0.47 & 0.32 & 0.47 & 0.32 \\
gemini                      & 0.47 & 0.43 & 0.47 & \textbf{0.43} & 0.47 & 0.43 & 0.47 & 0.43 & 0.47 & 0.43 & 0.45 & 0.41 \\
multilingual-e5-large       & 0.49 & 0.42 & 0.49 & 0.42 & 0.49 & 0.42 & 0.49 & 0.42 & 0.49 & 0.42 & 0.49 & \textbf{0.42} \\
mistral-embed               & 0.51 & \textbf{0.44} & 0.51 & 0.41 & 0.50 & \textbf{0.43} & 0.50 & \textbf{0.43} & 0.50 & 0.41 & \textbf{0.51} & 0.41 \\
text-embedding-3-small      & 0.46 & 0.41 & 0.47 & 0.40 & 0.46 & 0.41 & 0.46 & 0.41 & 0.46 & \textbf{0.43} & 0.45 & 0.42 \\
Solon-embeddings-large-0.1  & 0.50 & 0.39 & 0.50 & 0.39 & 0.50 & 0.39 & 0.50 & 0.39 & 0.50 & 0.39 & 0.50 & 0.39 \\
LaBSE                       & 0.46 & 0.38 & 0.46 & 0.38 & 0.46 & 0.38 & 0.46 & 0.38 & 0.46 & 0.38 & 0.46 & 0.38 \\
paraphrase-multilingual-MiniLM-L12-v2 & 0.37 & 0.20 & 0.37 & 0.20 & 0.37 & 0.20 & 0.37 & 0.20 & 0.37 & 0.20 & 0.37 & 0.20 \\
paraphrase-multilingual-mpnet-base-v2 & 0.41 & 0.26 & 0.41 & 0.26 & 0.41 & 0.26 & 0.41 & 0.26 & 0.41 & 0.26 & 0.41 & 0.26 \\
snowflake-arctic-embed-l-v2.0 & 0.46 & 0.40 & 0.46 & 0.40 & 0.46 & 0.40 & 0.46 & 0.40 & 0.46 & 0.40 & 0.46 & 0.40 \\
\midrule
\textbf{Mean}               & 0.46 & 0.37 & 0.47 & 0.36 & 0.46 & 0.37 & 0.46 & 0.37 & 0.46 & 0.36 & 0.46 & 0.36 \\
\midrule
\addlinespace[0.6ex]
\multicolumn{13}{c}{\textbf{OPTICS}}\\
\midrule
bge-m3                      & \textbf{0.41} & 0.33 & \textbf{0.41} & 0.33 & \textbf{0.41} & 0.33 & \textbf{0.41} & 0.33 & \textbf{0.41} & 0.33 & \textbf{0.41} & 0.33 \\
sentence-camembert-base     & 0.34 & 0.24 & 0.34 & 0.24 & 0.34 & 0.24 & 0.34 & 0.24 & 0.34 & 0.24 & 0.34 & 0.24 \\
gemini                      & 0.36 & 0.33 & 0.36 & 0.33 & 0.36 & 0.33 & 0.36 & 0.33 & 0.36 & 0.33 & 0.36 & 0.33 \\
multilingual-e5-large       & 0.38 & 0.32 & 0.38 & 0.32 & 0.38 & 0.32 & 0.38 & 0.32 & 0.38 & 0.32 & 0.38 & 0.32 \\
mistral-embed               & 0.37 & 0.33 & 0.37 & 0.33 & 0.39 & \textbf{0.35} & 0.38 & 0.33 & 0.37 & 0.33 & 0.38 & 0.33 \\
text-embedding-3-small      & 0.38 & 0.32 & 0.37 & 0.31 & 0.37 & 0.30 & 0.37 & 0.31 & 0.37 & 0.32 & 0.37 & 0.29 \\
Solon-embeddings-large-0.1  & 0.37 & 0.29 & 0.37 & 0.29 & 0.37 & 0.29 & 0.37 & 0.29 & 0.37 & 0.29 & 0.37 & 0.29 \\
LaBSE                       & 0.38 & 0.28 & 0.38 & 0.28 & 0.38 & 0.28 & 0.38 & 0.28 & 0.38 & 0.28 & 0.38 & 0.28 \\
paraphrase-multilingual-MiniLM-L12-v2 & 0.35 & 0.26 & 0.35 & 0.26 & 0.35 & 0.26 & 0.35 & 0.26 & 0.35 & 0.26 & 0.35 & 0.26 \\
paraphrase-multilingual-mpnet-base-v2 & 0.35 & 0.31 & 0.35 & 0.31 & 0.35 & 0.31 & 0.35 & 0.31 & 0.35 & 0.31 & 0.35 & 0.31 \\
snowflake-arctic-embed-l-v2.0 & 0.36 & \textbf{0.34} & 0.36 & \textbf{0.34} & 0.36 & 0.34 & 0.36 & \textbf{0.34} & 0.36 & \textbf{0.34} & 0.36 & \textbf{0.34} \\
\midrule
\textbf{Mean}               & 0.37 & 0.30 & 0.37 & 0.30 & 0.37 & 0.30 & 0.37 & 0.30 & 0.37 & 0.30 & 0.37 & 0.30 \\
\bottomrule
\end{tabular}
\caption{Per-model Fleiss' $\kappa$ for agreement across \texttt{UMAP} dimensionalities, shown for alignment thresholds $\ta \in [0.2,0.7]$.}
\label{tab:interdim_kappa_toa}
\end{table*}

\end{document}